\documentclass[10pt,twocolumn,letterpaper]{article}

\usepackage{iccv}
\usepackage{times}
\usepackage{epsfig}
\usepackage{graphicx}
\usepackage{amsmath}
\usepackage{amssymb}
\usepackage{booktabs}
\usepackage{algorithmic}
\usepackage{algorithm}
\usepackage{blindtext}
\usepackage{caption}
\usepackage{dsfont}
\usepackage{bbm}
\usepackage{bbding}
\usepackage[para]{footmisc}
\usepackage[square,comma,numbers]{natbib}

\usepackage{amssymb}
\usepackage[table]{xcolor}
\usepackage[pagebackref=true,breaklinks=true,letterpaper=true,colorlinks,bookmarks=false]{hyperref}

\iccvfinalcopy 

\def\iccvPaperID{****} 

\begin{document}

\title{DC-ControlNet: Decoupling Inter- and Intra-Element Conditions in \\ 
Image Generation with Diffusion Models}

\def\spaces{~~~~~}
\author{Hongji Yang\footnotemark[2],\spaces Wencheng Han\textsuperscript{\footnotemark[2]},\spaces Yucheng Zhou,\spaces Jianbing Shen\footnotemark[1]\\\\
SKL-IOTSC, CIS, University of Macau \\ 
{\tt\small yc47942@um.edu.mo, wencheng256@gmail.com}\\ {\tt\small yucheng.zhou@connect.um.edu.mo, jianbingshen@um.edu.mo}
}



\ificcvfinal\thispagestyle{empty}\fi

\twocolumn[{%
\renewcommand\twocolumn[1][]{#1}%
\maketitle

\begin{center}
    \centering
    \captionsetup{type=figure}
    \includegraphics[width=1.00\textwidth]{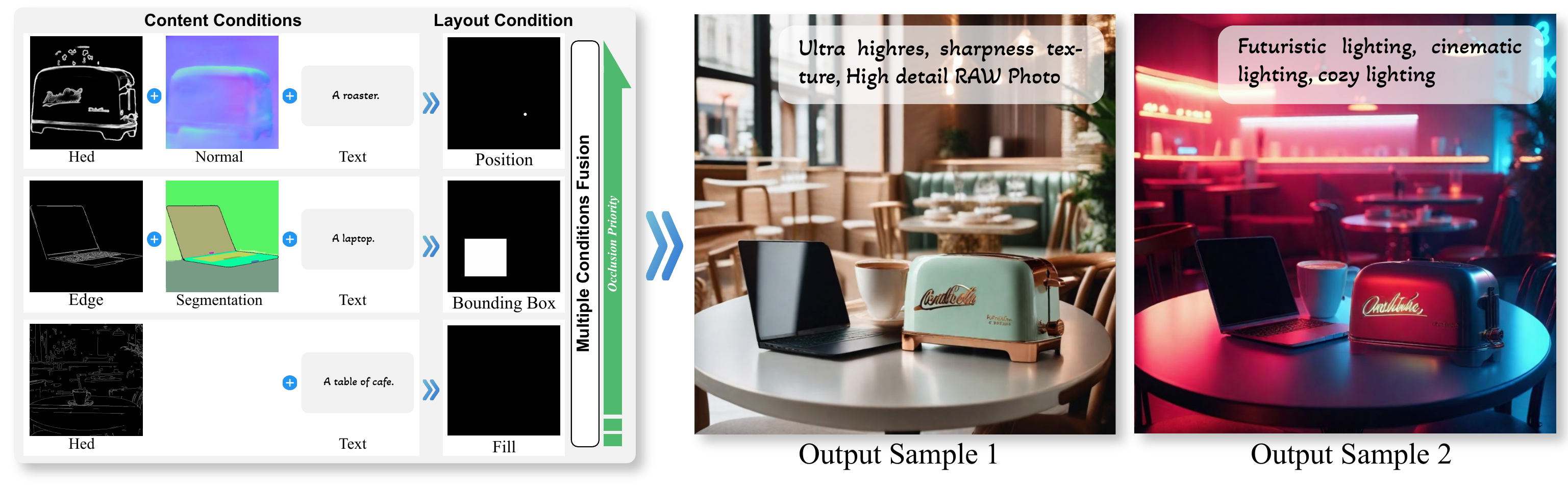}
    \captionof{figure}{Controllable generation using the proposed \textbf{DC-ControlNet} under various conditions. In our pipeline, each element can be controlled by decoupled attributes, while multiple elements can be naturally fused within a given order, enabling occlusion-aware and flexible customized generation. Results are $1024\times1024$.}
    \label{fig1}
\end{center}
}]
\renewcommand{\thefootnote}{\fnsymbol{footnote}} 
\footnotetext[2]{Equal contribution} 
\footnotetext[1]{Corresponding author}

\begin{abstract}
In this paper, we introduce DC (Decouple)-ControlNet, a highly flexible and precisely controllable framework for multi-condition image generation. The core idea behind DC-ControlNet is to decouple control conditions, transforming global control into a hierarchical system that integrates distinct elements, contents, and layouts. This enables users to mix these individual conditions with greater flexibility, leading to more efficient and accurate image generation control.
Previous ControlNet-based models rely solely on global conditions, which affect the entire image and lack the ability of element- or region-specific control. This limitation reduces flexibility and can cause condition misunderstandings in multi-conditional image generation. To address these challenges, we propose both intra-element and inter-element Controllers in DC-ControlNet. The Intra-Element Controller handles different types of control signals within individual elements, accurately describing the content and layout characteristics of the object. For interactions between elements, we introduce the Inter-Element Controller, which accurately handles multi-element interactions and occlusion based on user-defined relationships.
Extensive evaluations show that DC-ControlNet significantly outperforms existing ControlNet models and Layout-to-Image generative models in terms of control flexibility and precision in multi-condition control. 

\noindent\textit{Our project website:} \url{https://um-lab.github.io/DC-ControlNet/}.
\end{abstract}



\section{Introduction}
\begin{figure}
    \centering
    \includegraphics[width=1.0\linewidth]{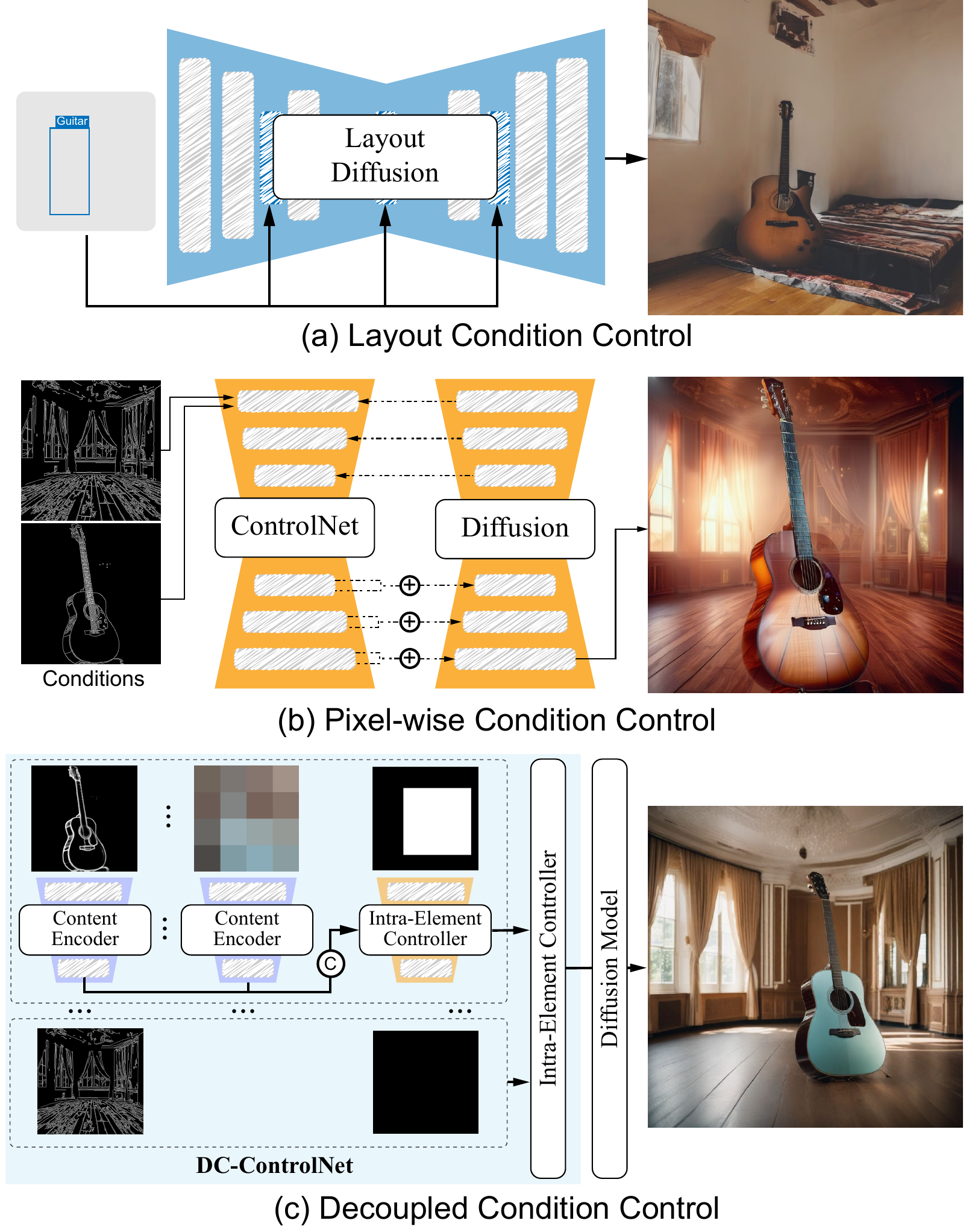}
    \caption{\textbf{Comparison between existing methods and ours}. (a) \textbf{layout-to-image models}~\cite{zheng2023layoutdiffusion,cheng2024hico} misalign the shape of the user-specified ``guitar'' accurately. Meanwhile, (b) existing pixel-wise conditions \textbf{ControlNet}~\cite{zhang2023adding,li2025controlnet} cause unsatisfactory textures and layouts in conflict areas when fusing multiple conditions. (c) \textbf{DC-ControlNet} offers a more flexible and precise solution for integrating multi-condition control.
    }
    \label{fig:shortcoming}
    \vspace{-2mm}
\end{figure}

Condition control plays a crucial role in image generation~\cite{li2024blip,li2024photomaker,mou2024t2i,shi2024instantbooth,wei2023elite,xiao2024fastcomposer}, requiring the output images to closely align with the designer's specifications, including style~\cite{zhang2023inversion,everaert2023diffusion}, content~\cite{zhang2023adding,controlnetplus}, layout~\cite{zheng2023layoutdiffusion,zhao2020layout2image,cheng2024hico} and other attributes. Most image generation methods~\cite{esser2024scaling,flux,sdxl,rombach2022high} rely on text prompts as the primary mechanism for condition control. Due to the expressive and versatile nature of natural language, text prompts integrate multiple aspects of the conditions into a single input, allowing for holistic control over the generated image.
However, the layout-to-image generative model struggles to control the specific shape of an element in an image, as shown in Fig.~\ref{fig:shortcoming}~(a). It cannot control the detail color and texture of the ``guitar''.



To achieve precise detail control, Zhang~\etal~\cite{zhang2023adding} introduced ControlNet, which integrates pixel-wise control conditions into the image generation process. This approach establishes a clear relationship between the output and the conditions, allowing for accurate control over the output by specifying detailed conditions and overcoming the limitations of purely text-based control. However, ControlNet only supports global conditions, resulting in each condition image influencing the entire image and does not allow for element-specific control. This limitation presents significant challenges for multi-condition image generation.  As shown in Fig.\ref{fig:shortcoming}~(b), when multiple conditions are provided to describe different parts of the image, ControlNet struggles to accurately understand the relationships between these conditions. It cannot determine whether the conditions refer to the same element or different elements, resulting in incorrect artifacts.

To address these challenges, this paper introduces a novel approach that decouples control conditions by breaking down global control into a hierarchical integration of individual contents, and layouts, as shown in Fig.\ref{fig:shortcoming}~(c). Using simple interactions to combine multiple conditions, users can generate results that align with their intentions, achieving flexible and precise control.
Specifically, we first decompose the global image control into independent elements, such as different foreground objects and the background. Each element is controlled individually, and an Inter-Element Controller is then used to logically fuse these elements. This module accounts for inter-element relationship processing, such as lighting variations and occlusions, and integrates them based on the user design. This decoupling enables users to prepare conditions for each element independently, greatly enhancing flexibility.

Within each element, we further decouple the control into two types of conditions: Content conditions and Layout conditions. Content conditions focus on the intrinsic properties of the image, such as semantics, edges, and colors, while Layout conditions address spatial properties, such as position and size. By decoupling different conditions within each element, we further enhance the flexibility and accuracy of control. Each condition only focuses on specific aspects of the element, without the need to interact with other conditions or affect other elements, thereby resolving the issue of multi-condition misunderstandings.
As shown in Fig.\ref{fig:shortcoming}(c), the color condition specifies only the desired color for an element, without considering its detailed shape. 
To demonstrate the capability of DC-ControlNet in handling complex multi-condition image generation, we propose a new dataset and the corresponding benchmark, named Decoupled Multi-Condition (DMC-120k). 
This dataset includes a total of 120,000 diverse images with multiple conditions, and it will be made publicly available.
The main contributions are summarized as follows:
\begin{itemize}
\item We introduce DC-ControlNet, which decouples global conditions control into a hierarchical integration of individual contents and layouts, facilitating multi-condition image generation.
\item We propose the Intra-Element Controller, enabling independent control of each attribute within an element, without coupling different aspects.
\item {We present the Inter-Element Controller to effectively handle interactions between multiple individual elements, enabling accurate multi-element conditional image generation.} 
\item We construct a novel dataset, DMC-120k, which contains a wide range of samples with multi-conditions. Experiments based on this dataset show that our method significantly outperforms previous control methods in terms of flexibility and precision in multi-condition image generation.
\end{itemize}

\section{Related Work}
\noindent\textbf{Image Generation by Diffusion Models.}
Diffusion models~\cite{rombach2022high, ho2020denoising} have proven to be highly effective in a variety of generative tasks, offering significant improvements in semantic relevance and image quality compared to GAN-based models~\cite{goodfellow2020generative}. Through iterative image denoising, diffusion models excel in generating high-quality images.
The Denoising Diffusion Probabilistic Model (DDPM)~\cite{ho2020denoising} introduced a probabilistic generative framework using a diffusion process. Building on this, DDIM~\cite{song2020denoising} and PLMS~\cite{liu2022pseudo} improved generation efficiency by utilizing an implicit process and a pseudo-likelihood function.
The Latent Diffusion Model (LDM)~\cite{rombach2022high} incorporated a Variational Autoencoder (VAE)~\cite{kingma2013auto} to map images into a latent space, and facilitated the generation of larger and higher-quality images.
Beyond the Unet-based architecture~\cite{ronneberger2015u}, many researcher also explored the integration of Transformer architectures into diffusion models for improved performance. Diffusion Transformer (DiT) models~\cite{chen2024pixartalpha, bao2023all, zhang2023adding, peebles2023scalable, chen2024pixart} have demonstrated improved scalability and effectiveness.

\noindent\textbf{Controllable Generation.} 
To introduce controllability, conditional diffusion models have been developed to guide the diffusion process using additional information, such as categories or text prompts. Models like Guided Diffusion~\cite{nichol2021glide} and Classifier-Free Guidance (CFG)~\cite{ho2021classifierfree} enable more flexible control by adjusting the strength of conditioning.
Going beyond implicit conditioning, ControlNet~\cite{zhang2023adding} introduced pixel-wise control through explicit mechanisms, using additional control signals like canny edges, depth maps, or OpenPose keypoints to guide the generation process. 
Similarly, T2I-Adapter~\cite{mou2024t2i} aligned internal knowledge in text-to-image models with external control signals, improving control precision.
IP-Adapter~\cite{ye2023ip} offered greater flexibility in controlling image generation by using the image prompt as a condition, allowing for the integration of multiple conditions simultaneously.
UniControlNet~\cite{zhao2023uni} and UniControl~\cite{qin2023unicontrol} introduced a union control framework for controllable condition-to-image with different conditions.
ControlNeXt~\cite{peng2024controlnext} employs Cross-Normalization to align the condition’s distribution with the main branch, facilitating faster convergence. ControlNet++~\cite{li2025controlnet} further refined this approach by incorporating reinforcement learning to better align the diffusion model with the given conditions.

Additionally, the layout has been used as a complementary control signal in text-to-image generation~\cite{karacan2016learning, reed2016learning,wang2022interactive}. 
Layout2Im~\cite{zhao2020layout2image} was the first to incorporate layout as a condition, defining it as a set of objects with bounding boxes and categories. 
LostGAN-v2~\cite{sun2021learning} employed a reconfigurable layout to control individual objects. 
GeoDiffusion~\cite{chen2024geodiffusion} introduced a geometric control approach, encoding location and object descriptions into pre-trained diffusion models. 
DetDiffusion~\cite{wang2024detdiffusion} proposed a perception-aware loss to improve the generation quality and controllability.
LayoutDiffusion~\cite{zheng2023layoutdiffusion} treated each image patch as a distinct object, achieving the integration of layout and image in a unified manner. 
HiCo~\cite{cheng2024hico} achieved spatial disentanglement by hierarchically modeling layouts.
However, these methods restrict layout information to the position defined by bounding boxes and the semantics specified by categories. As a result, they fall short of customizing the specific content within the boxes based on the given conditions.

\begin{figure}
    \centering
    \includegraphics[width=1\linewidth]{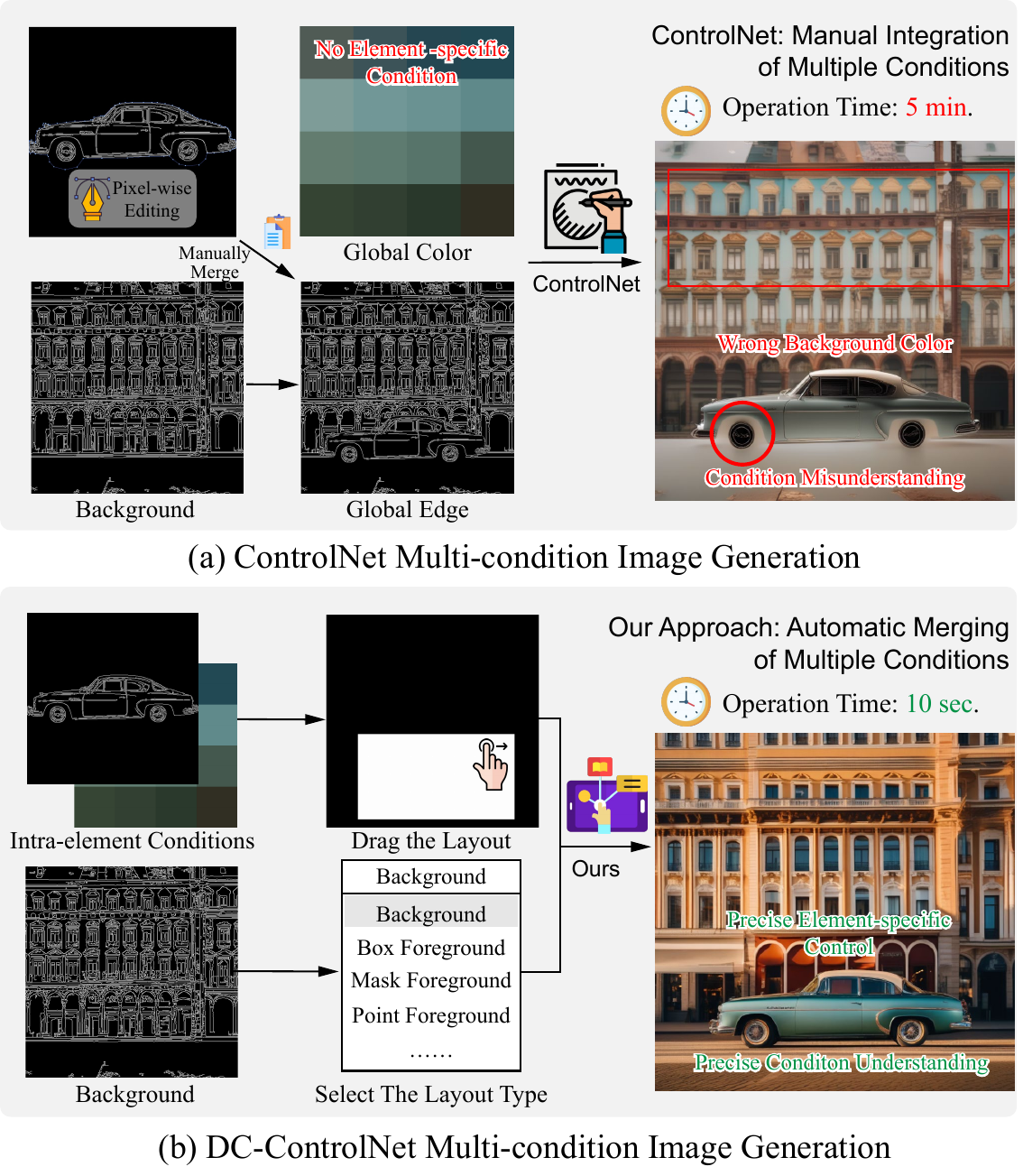}
    \vspace{-1mm}
    \caption{\textbf{Comparison of user interaction between (a) ControlNet~\cite{zhang2023adding} and (b) our {DC-ControlNet}}. Users can selectively identify and separately control specific elements, thus achieving an efficient controllable generation.}
    \label{fig:edit}
    \vspace{-6mm}
\end{figure}

\begin{figure*}
    \centering
    \includegraphics[width=1\linewidth]{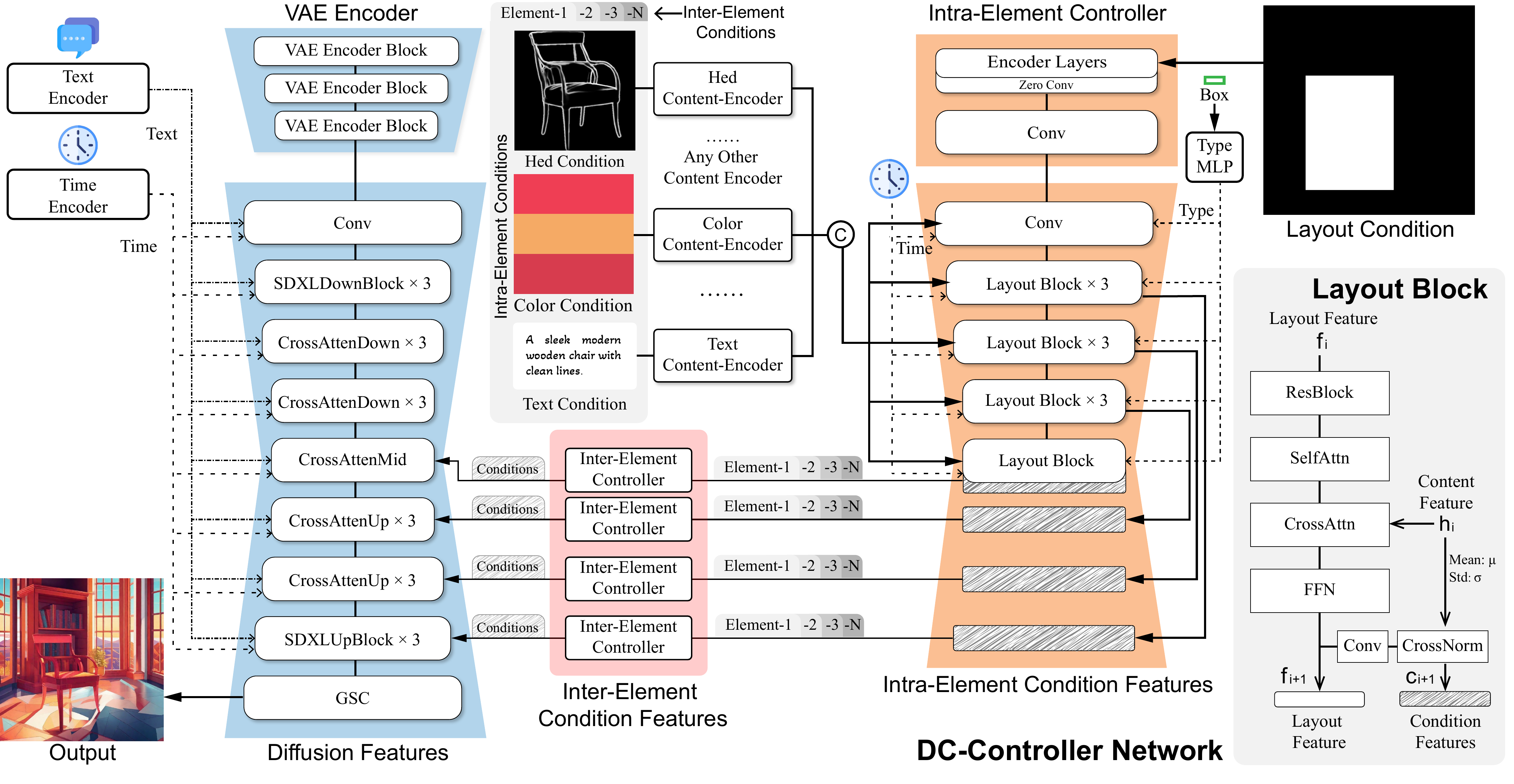}
    \caption{\textbf{The overall pipeline of our DC-ControlNet.} The purpose of our DC-ControlNet is to transform the output content features onto the target layout. An Intra-Element Controller consists of 10 layout blocks, where each block receives a content feature and layout feature. To enable model controlling different types of layout \{dot, box, mask\}, we insert a layout-type embedding. Then, if multiple elements are involved, the condition features are further fused through the Inter-Element Controller to resolve conflicts.}
    \label{fig:over-pipeline}
    \vspace{-3mm}
\end{figure*}

\section{Methodology}
\subsection{Preliminary}
Diffusion Models (DMs) are powerful generative models that have shown remarkable success in image and video generation. By reversing the noise-adding process in latent space, latent diffusion models (LDMs) transform random noise into a target latent variable. The prediction for $z_{t}$ at time step $t$ is determined by $z_{t+1}$ and $t$. The forward adding noise process and reverse de-noise process can be represented as:
\begin{equation}
    q(z_{t}|z_{t-1})=\mathcal N(z_{t};\sqrt{1-\beta_{t}}z_{t-1}I)
\end{equation}
\begin{equation}
    p_{\theta}(z_{t-1}|z_{t}) = \mathcal N(z_{t-1};\mu_\theta{(z_t,t)},\sigma_t^2 I)
\end{equation}
where $\beta_t$ is a coefficient that controls the noise strength in step $t$ and $\theta$ represents the trained model for giving the prediction $\mu_\theta{(z_t,t)}$.
The loss function of diffusion models is the MSE loss that fits the noise $\epsilon$ by the given noise data $z_t$, timesteps $t$, and condition $c$.
\begin{equation}
    \mathcal L_{mse} = \Vert \epsilon - \epsilon_\theta(z_t, t, c) \Vert_2^2
\end{equation}

To provide detailed control, ControlNet~\cite{zhang2023adding} introduces pixel-wise conditions $c_{f}$ as an additional input. The loss function can be expressed as follows:
\begin{equation}
    \mathcal L_{mse} = \Vert \epsilon - \epsilon_\theta(z_t, t, c, c_{f}) \Vert_2^2
\end{equation}

\subsection{The Proposed Decoupled ControlNet}
The original ControlNet requires users to provide global conditions to control the generation of an entire image. This approach forces users to combine all elements, such as foreground objects, the background, and their relative positions, into one input, significantly reducing control flexibility. This can complicate the editing process, leading to confusion and potential artifacts, as discussed in the introduction. To address this limitation, our pipeline decouples the control conditions by breaking down the global conditions of ControlNet into a hierarchical organization that manages distinct elements and their individual contents and layouts. This approach enhances the flexibility of user editing, allowing for precise, independent control over each element and its internal conditions.

Fig.~\ref{fig:edit} compares our method to ControlNet. In this case, users want to blend multiple conditions, such as the edge of the background and foreground vehicle and the color of the vehicle. ControlNet requires manual pixel-wise editing of the condition image, followed by merging it into the global condition. This process is time-consuming and prone to causing misunderstandings in the model. Additionally, ControlNet does not support element-specific conditions, which limits the user’s ability to control features like vehicle color without affecting the background. In contrast, our method allows for precise condition setting with simple point-and-drag interactions, supporting element-specific control to accurately achieve the user’s desired foreground color adjustment.

As illustrated in Fig.~\ref{fig:over-pipeline}, our pipeline consists of four main components: the Diffusion Model, Content Encoder, Intra-Element Controller, and Inter-Element Controller. The Content Encoder is responsible for encoding the content conditions of each element, capturing different aspects within the element. The Intra-Element Controller integrates the content conditions and their corresponding layout for each individual element. The Inter-Element Controller integrates the conditions of multiple elements, merging them into a final unified condition, which is then passed to the Diffusion Model to generate the desired image.
\begin{figure}
    \centering
\includegraphics[width=1\linewidth]{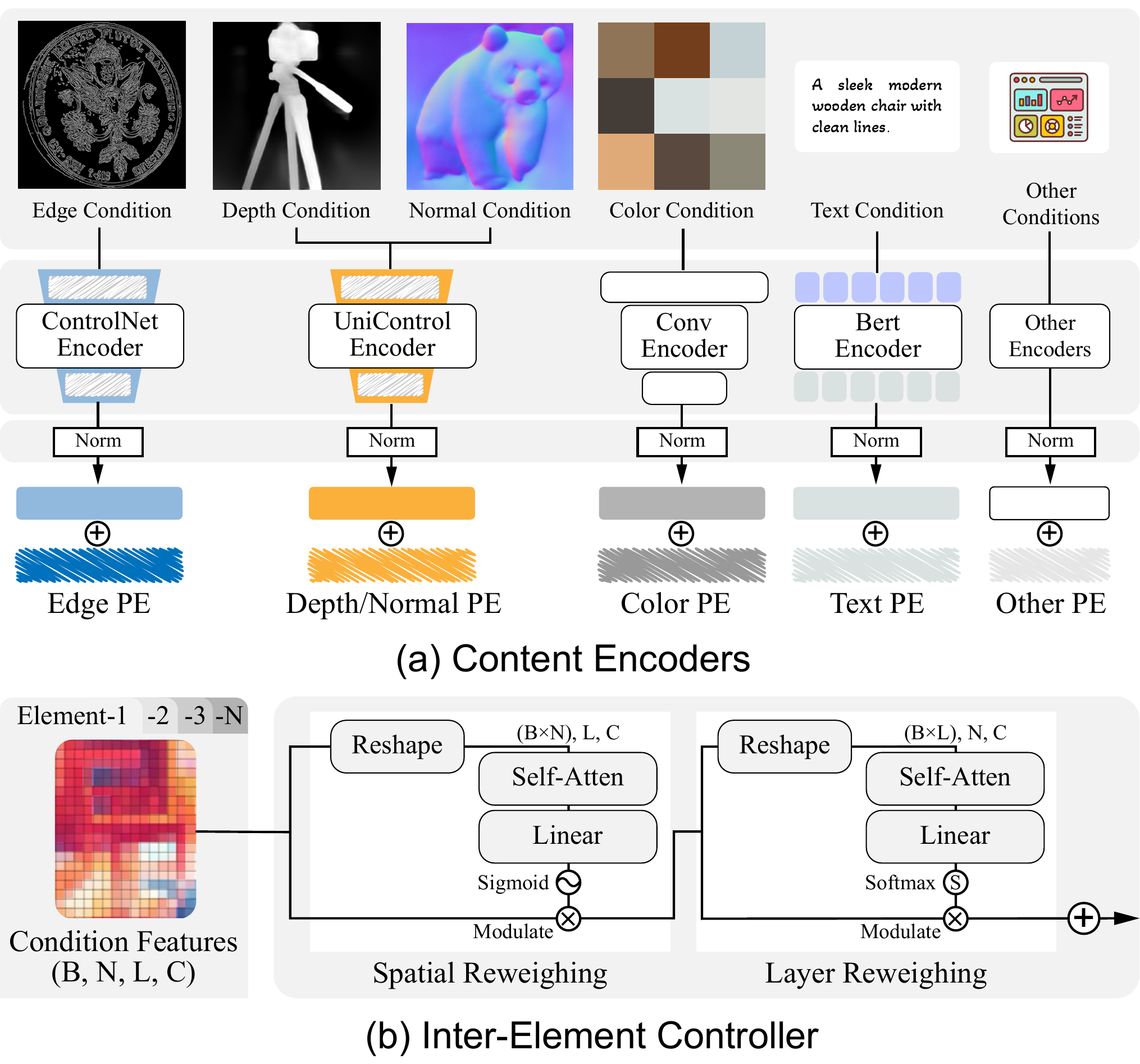}
    \caption{\textbf{The component of our DC-ControlNet.} (a) \textbf{Different Content Encoders} in our intra-element controller to control the generated images across various aspects; (b) \textbf{The structure of the Inter-Element Controller} for adjusting the order of different elements.}
    \label{fig:content-encoders-inter-element-controller}
    \vspace{-4mm}
\end{figure}

\subsection{Intra-Element Controller}
In practical creative tasks, users often rely on various control conditions to guide different aspects of target generation. 
For instance, edge conditions can regulate image details and textures, while depth and normal conditions define geometric properties, and color conditions control surface colors.
However, since each condition focuses on a distinct aspect, it is challenging for a unified encoder to accurately encode all these conditions. To address this, our pipeline supports a wide range of content encoders, allowing users to integrate the appropriate encoder into the proposed DC-ControlNet framework.

Fig.~\ref{fig:content-encoders-inter-element-controller}~(a) illustrates several content encoders employed in our method, including the ControlNet~\cite{zhang2023adding} Encoder for edge information, the ControlNetPlus~\cite{controlnetplus} Encoder for geometric attributes, the ConvEncoder, a simple convolutional neural network for encoding color information, and BERT~\cite{devlin2018bert} for text information. 
Additionally, we assign the corresponding positional embeddings to each type of condition. Finally, we concatenate the multi-condition contents along the length dimension before passing features into the Intra-Element Controller.
%


Once the intra-element content is defined by content encoders, the next step is to consider how to inject these contents into the given layout.
For the layout image, we define it to be a dot map, box map, or mask map, ranging from weak to strong layout control, which can specify where the element is located (i.e., dot map), the approximate size and location of the element (i.e., box map), or the exact shape and location of the element (i.e., mask map).

The core idea of our Intra-Element Controller is to extract pixel-wise spatial relationships from the layout condition and then inject the content feature into the corresponding areas in a learnable manner. During this injection process, the Intra-Element Controller considers the relative relationships between elements and their associations with the environment, ensuring that the overall layout is accurate and harmonious.
{{Specifically, for a layout representation of an element, we first embed it into a layout embedding \( f_{0} \in \mathbbm{R}^{b\times h \times w \times c} \) through several convolution layers and then add it to the  $x$  in the diffusion UNet. Next, we apply a control-type embedding to indicate the type of content condition, which is injected into the ResBlock in a manner similar to time embedding.}}

Considering the content feature $\{h_{1}, h_{2}, \cdots, h_{i}\}$ extracted from Content Encoder, our goal is to equip these features with layout information and output them at their original resolution, so that they can still be efficiently injected into the Unet as ControlNet features.
As shown in Fig.~\ref{fig:over-pipeline}, the controller contains multiple fusion blocks to extract the layout features at different resolutions and inject content features into them. 
Specifically, our layout block contains a ResBlock and a Transformer block, where the ResBlock receives time embedding and layout control-type embedding, and the Transformer block employs cross-attention with the content feature as key and value, followed by a convolution layer for output.

Since layout and content are not pixel-aligned, we apply a differential positioning strategy in cross-attention to give the condition a different position embedding than the query to better amplify this difference. We employ the attention with RoPE~\cite{su2024roformer} and add a fixed offset vector $\Delta$~\cite{tan2024ominicontrol} to the positional embedding of the key. 
\begin{equation}
    {(i,j)}_{\text{content}} = {(i,j)}_{\text{layout}} + \Delta
\end{equation}
where $(i,j)$ denotes the position of the image token.

After the ResBlock and transformer block, we obtain the content features with layout information for Unet. Unfortunately, after this fusion module, Layout-ControlNet introduces distortion in the distribution of the original content feature, which causes slower convergence. Therefore, we apply the Cross-Normalization to recover the output feature with the mean and standard deviation of the original content features. In this way, the output features are consistent in distribution with the original ControlNet's features, avoiding sudden collapses due to large distribution differences or the extra cost of using zero-conv to slowly align the Unet.
\begin{equation}
    \begin{aligned}
    & f_{i}^{'} = f_{i}^{'} + \text{SelfAttn}(f_{i}^{'}) \\
    & f_{i}^{'} = f_{i}^{'} + \text{CrossAttn}(f_{i}^{'}, h_{i}) \\
    & f_{i+1} = f_{i}^{'} + \text{FFN}(f_{i}^{'}) \\
    & h_{i}^{'} = \text{Conv}(f_{i+1}) \\
    & h_{i}^{'} = \left( \frac{h_{i}^{'} - \mu_{h_{i}^{'}}}{\sigma_{h_{i}^{'}}} \right) \cdot \sigma_{h_{i}} + \mu_{h_{i}} \\
    \end{aligned}
\end{equation}

In this way, the layout information can be fully utilized to enable controllable generation of the specified layout. 
Moreover, its output remains fully compatible with the original ControlNet features to maintain its scalability.

\subsection{Inter-Element Controller}
The original ControlNet enables multi-type conditions controllable generation by adding the feature directly.
However, this strategy results in an unreasonable feature fusion, as individual feature maps spatially share a single strength, leading to unnecessary artifacts.
%


When dealing with occlusion problems, such as when ``object A is in front of object B'', simply adjusting the scale does not achieve such a fusion. This is because the model cannot infer the relative positioning of elements based on scale alone. However, our approach enables the fusion of different conditions for multiple elements, even in the case of conflicts.

\noindent{{\textbf{Order embedding.}}
We define this occlusion problem as a problem of layer ordering. Specifically, in overlapping regions, features in {upper layers} should have a higher priority and should not be obscured by features at {lower layers.} When inference, \textit{Object A is in front of object B} can be easily realized by simply adjusting the layer order. 
We treat $L$ elements and concatenate them according to their temporal order to form an input similar to the video latent features. Similarly, to enable the model to perceive the layer order, we provide it with an order embedding to determine the order relationship of these elements. 

Considering a set of $L$ elements $\{x_{1}, x_{2}, ..., x_{l}\}$, we first perform an order sorting of the occlusion relationships and then concatenate them into a 4-d input:
\begin{equation}
    x^{'} = [\text{Sorted}(x_0,x_1, ..., x_{l})], x^{'} \in \mathbbm{R}^{B\times L\times N \times C}
\end{equation}
where the $B$, $L$, $N$, and $C$ denote the batch size, layer dimension, spatial dimension, and channel dimension, respectively. 

After assigning the order embeddings, we obtain a multi-condition representation that describes both the order and spatial relationships. In our implementation, since we use RoPE, the order embedding is applied before the attention, rather than directly adding to the features.

%
\noindent{{\textbf{Spatial and Layer Reweighing Transformer.}}
%
To accurately distinguish the position each element occupies in both the spatial and layer dimensions, we employ spatial and layer transformers to predict spatial and layer weights for reweighing. 
The structure of the fusion module is shown in Fig.~\ref{fig:content-encoders-inter-element-controller}~(b). 
{Specifically, we apply the spatial transformer and layer transformer to predict the weight of feature at the spatial dimension and layer dimension, respectively.} These transformers are used solely to predict the weights, rather than directly modifying the features. Before the spatial/layer transformer, we first merge the irrelevant dimensions and then feed them into the corresponding transformer. In this step, we apply RoPE to encode the positions of the spatial or layer dimensions, helping the model to perceive the relative spatial positions or the order of elements. The process of spatial transformer and layer transformer can be expressed in Algorithm~\ref{algo:srt}. and Algorithm~\ref{algo:crt}, respectively.



\vspace{-2mm}
\begin{algorithm}[H]
    \caption{Spatial reweighing Transformer}
    \renewcommand{\algorithmicrequire}{\textbf{Input:}}
    \renewcommand{\algorithmicensure}{\textbf{Output:}}
    \label{algo:srt}
    \begin{algorithmic}[1]
        \REQUIRE The representation $x \in \mathbbm{R}^{B\times L\times N \times C} $  
        \ENSURE The spatial-reweighed $x \in \mathbbm{R}^{B\times L\times N \times C} $
        
        $x = \text{rearrange}(x, \:b\: l\: \:n \:c \rightarrow (\:b \:l\:) \:n \:c)$

        $x^{'} = \text{LayerNorm}(x)$

        $q = W_{q}\cdot x^{'} , k = W_{k}\cdot x^{'}  , v = W_{v}\cdot x^{'}   $

        $q^{'} = q\cdot R(\theta), k^{'} = k\cdot  R(\theta)$ \COMMENT{2d pos embedding}

        $x^{'} = x + \text{Attention}(q^{'}, k^{'}, v)$

        $x^{'} = x^{'} + \text{FFN}(\text{LayerNorm}(x^{'}))$

        $w_{\text{spatial}} = \text{Sigmoid}(\text{Linear}(x^{'}))$ \COMMENT{zero-init linear}

        $x = x * w_{\text{spatial}}$ \COMMENT{reweighing}
        
        $x = \text{rearrange}(x, (b\: l\:) \:n \:c \rightarrow b \:l \:n \:c)$
        \end{algorithmic}
\end{algorithm}
\vspace{-3mm}

\begin{algorithm}[H]
    \caption{Layer reweighing Transformer}
    \renewcommand{\algorithmicrequire}{\textbf{Input:}}
    \renewcommand{\algorithmicensure}{\textbf{Output:}}
    \label{algo:crt}
    \begin{algorithmic}[1]
        \REQUIRE The representation $x \in \mathbbm{R}^{B\times L\times N \times C} $  
        \ENSURE The layer-reweighed $x \in \mathbbm{R}^{B\times L\times N \times C} $
        
        $x = \text{rearrange}(x, \:b\: l\: \:n \:c \rightarrow (\:b \:n\:) \:l \:c)$

        $x^{'} = \text{LayerNorm}(x)$

        $q = W_{q}\cdot x^{'} , k = W_{k}\cdot x^{'}  , v = W_{v}\cdot x^{'}   $

        $q^{'} = q\cdot R(\theta), k^{'} = k\cdot  R(\theta)$ \COMMENT{1d order embedding}

        $x^{'} = x + \text{Attention}(q^{'}, k^{'}, v)$

        $x^{'} = x^{'} + \text{FFN}(\text{LayerNorm}(x^{'}))$

        $w_{\text{layer}} = \text{Softmax}(\text{Linear}(x^{'}))$ 

        $x = x * w_{\text{layer}}$ \COMMENT{reweighing}
        
        $x = \text{rearrange}(x, (b\: n\:) \:l \:c \rightarrow b \:l \:n \:c)$
        \end{algorithmic}
\end{algorithm}
\vspace{-3mm}
Here, b, l, n, c represent the batch size, number of layers, number of image tokens and channel dimensions. Note that in spatial reweighing transformer, we apply 2d positional embedding. For simplicity of expression, we unify the 2d positional embedding and 1d order embedding under the notation $R(\theta)$.
        
        
The output is then passed through a linear layer and an activate function to obtain the spatial or layer weights. 
The spatial weights serve to emphasize the main content of each element, while the layer weights highlight which element plays a significant role for each image token. 
{As the relationships between different layers of the same pixel are mutually exclusive, we apply softmax to obtain the final weight, which distinguishes our method from the spatial reweighting transformer. When obtaining spatial weights, we use a zero-initialized linear layer followed by a sigmoid activation. With these improvements, we achieve precise control by reweighting the features along both the spatial and layer dimensions. Furthermore, this reweighting approach prevents the model from taking shortcuts by directly modifying the content of specific elements. The final output is obtained by summing along the layer dimension.}

\subsection{Loss Function}


To further facilitate the model to learn the distribution of the foreground, we follow~\cite{chen2024geodiffusion}, employing the re-weighting strategy to assign a greater weight to the foreground. Specifically, we set the weight of the foreground region as inverse proportion to its area relative to the total latent area.
\begin{equation}
m_{i,j} = 
\begin{cases} 
1, & (i,j) \in \text{background} \\
\frac{\text{Aera}_{\text{total}}}{\text{Aera}_{\text{foreground}}}, & (i,j) \in \text{foreground}
\end{cases} 
\end{equation}

Here, the weight of the pixel $(i,j)$ is assigned based on the area of the foreground region, which is set equal to that of the entire image. This implies that the optimization objective for the foreground region is equivalent to that of the background. The smaller the foreground region, the larger its weight, ensuring that the target is not ignored. Note that the loss reverts to the original one when the area of the foreground region equals the total image area. 
Second, a feature-wise L1 loss $L_{transform}$ supervises the spatial transformation of features in the intra-element controller by comparing transformed features $ h^{'}_{i}$ with target feature $\hat{h}^{'}_{i}$ from the ControlNet. Both loss is weighted by $m$.
The final optimization objective can be defined as follows:
\begin{equation}
    \mathcal L_{transform} = \Vert h^{'}_{i} - \hat{h}^{'}_{i} \Vert_1
    \label{l_transform}
\end{equation}
\begin{equation}
    \mathcal L_{total} = \mathcal L_{mse} \cdot m + \lambda\mathcal L_{transform} \cdot m
\end{equation}
where $\lambda$ is a coefficient to balance the weight of the loss.

\begin{figure}[t]
    \centering
    \includegraphics[width=1\linewidth]{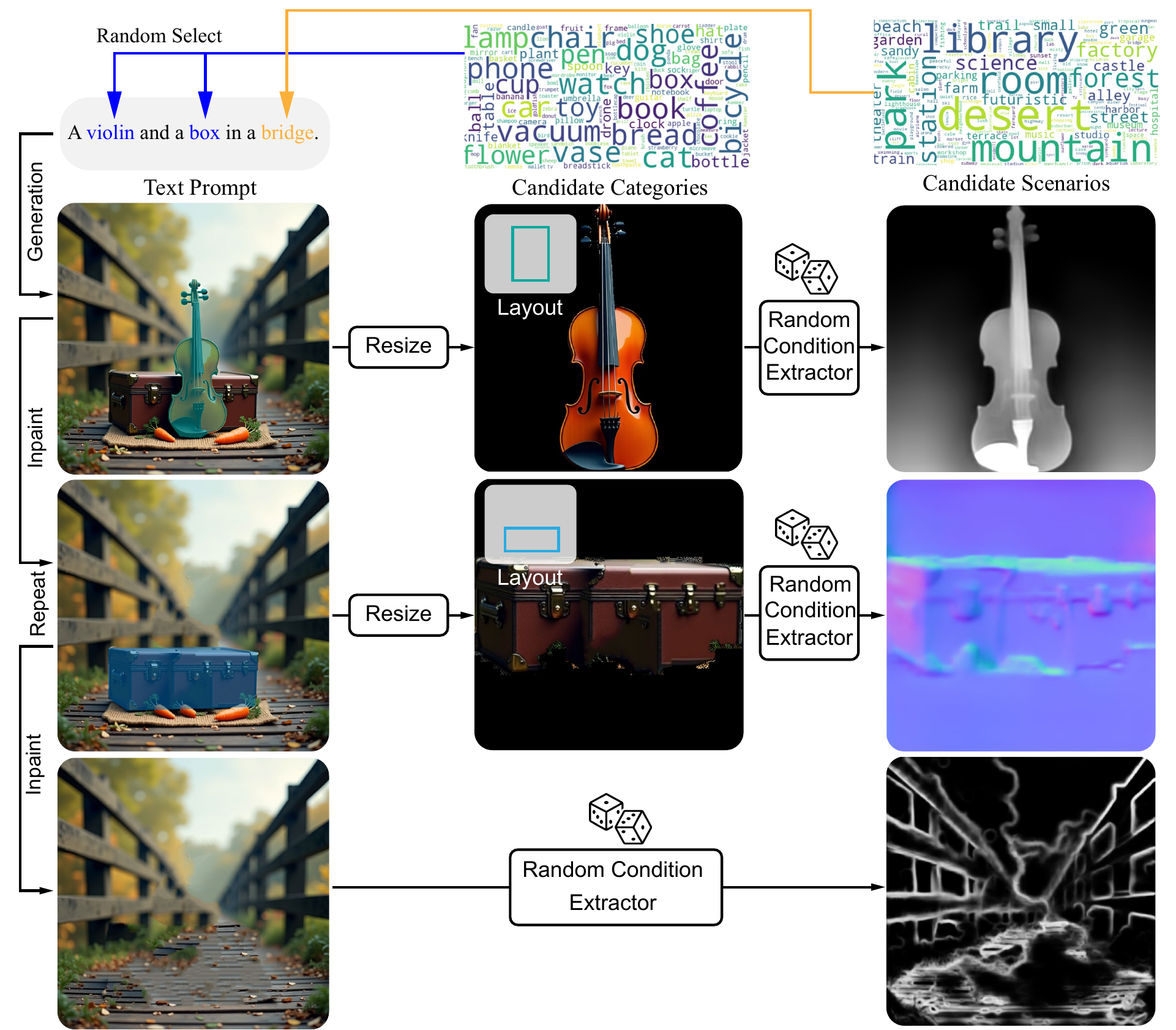}
    \caption{\textbf{The proposed DMC-120k dataset}. We first generate the multi-element image with random sampled categories and scenarios and inpaint the elements in order. The conditions are then obtained by corresponding condition extractors.}
    \label{fig:dataset}
    \vspace{-2mm}
\end{figure}

\begin{figure*}
    \centering
    \includegraphics[width=1\linewidth]{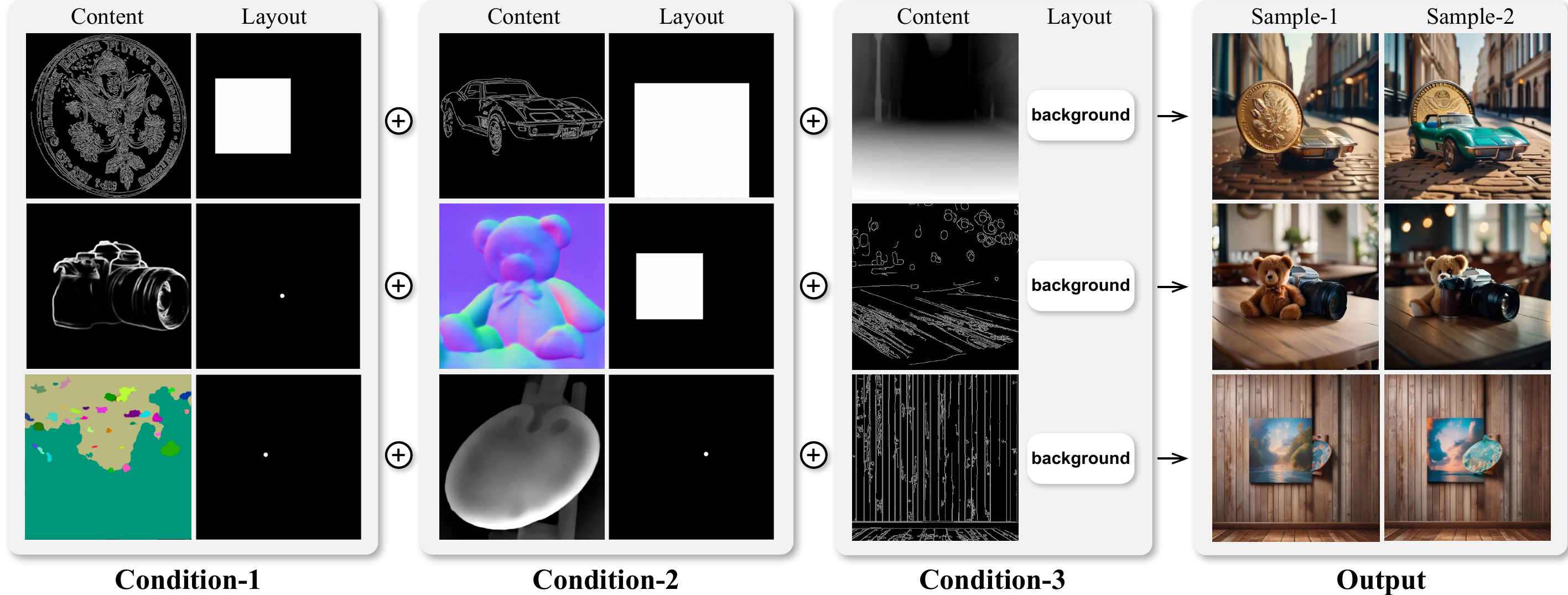}
    \caption{\textbf{The results of our DC-ControlNet}. Each element can generate pixel-aligned output through given content conditions, as well as the layout conditions. Besides, in overlapping regions, users can resolve conflicts by adjusting the layer order and leveraging our layer fusion module to achieve occlusion-aware generation.}
    \vspace{-1mm}
    \label{fig:total_exp}
\end{figure*}

\section{The Proposed DMC-120k dataset}
\label{sec:dataset}
%
{To fully leverage the potential of our method for multi-conditional image generation, we introduce a new high-quality dataset, DMC-120k, along with a new benchmark.
The core innovation behind this dataset is the idea of decoupling a complex image containing multiple elements into several independent conditions. These decoupled conditions are then used as training data to guide a model in reconstructing the original image.
Specifically, the DMC-120k dataset comprises 120,000 samples, each containing at least two independent elements with their ${e_{\text{content}}, e_{\text{layout}}}$, and a final target image, $I_{\text{trg}}$. The dataset is split into a training set of 120,000 samples and a test set of 1,000 samples. \textit{The entire dataset will be made publicly available.}
The total pipeline of dataset construction is shown in Fig.~\ref{fig:dataset}.

\noindent\textbf{Image Creation.} The first step is to generate complex images composed of multiple distinct elements. To achieve this, we define more than 100 different category labels, including animals, furniture, household items, and \textit{etc.}. We then construct prompts based on randomly selected categories and scenarios. These prompts are further optimized using ChatGPT to improve their suitability for the generation model and to enhance the variety of generated prompts. For each prompt, we use a random seed and generate high-resolution images at $1024 \times 1024$ pixels, employing the open-source SDXL~\cite{sdxl} and FLUX~\cite{flux} models to generate high-quality image data under various conditions.

\noindent\textbf{Element Decoupling.} Next, we use GroundingDino~\cite{liu2025grounding} to detect the location and shape of each object in the image. After obtaining the mask, we crop the foreground object and upsample it to $1024 \times 1024$ pixels, which is then used to extract the foreground conditions.
If two or more foreground element masks overlap, we label them as potentially occluding each other. To address occlusion, we first erase one of the masks and apply SDXL-Inpainting~\cite{sdxl} to inpaint the image. We then redetect the remaining objects to ensure they are intact. This process is repeated for all occluding elements until no occlusion remains. To extract the background conditions, we apply SDXL-Inpainting using the masks of all the foreground objects and inpaint the image.

\noindent\textbf{Condition generation.} The final step is to obtain the conditions. We achieve this by using various condition detectors for both foreground and background images. In our dataset, we provide detailed conditions such as canny, {HED}, depth, segmentation, and normal maps for content control. Additionally, we include dot maps, box maps, and mask maps as layout control conditions.
}

\section{Experiment}
\subsection{Experimental Setup}
Our DC-ControlNet is based on SDXL~\cite{sdxl} to achieve controllable generation. All experiments are conducted on eight A100 GPUs with mixed precision training. Our training is divided into three stages for training Union-ControlNet, Intra-Element Controller and Intra-Element Controller, respectively.
For Union-ControlNet, we use the following eight conditions: canny, HED, depth, segmentation map, normal, dot map, mask map, and box map for training. The model is trained using AdamW optimizer with a fixed learning rate of 1e-4 for 50,000 steps. 
For Intra-Element Controller, we employ the AdamW optimizer with a fixed learning rate of 1e-4 for 50,000 steps with a batch size of 32 and training for about one day. The dropout rate of the prompt is set to 0.2 in training. All training images, including the target image and the conditioning images, are set to $1024\times1024$ in the model. 
For the Inter-Element Controller, we use the same hyperparameters for training as for the Intra-Element Controller. 

\begin{figure*}
    \centering
    \includegraphics[width=1\linewidth]{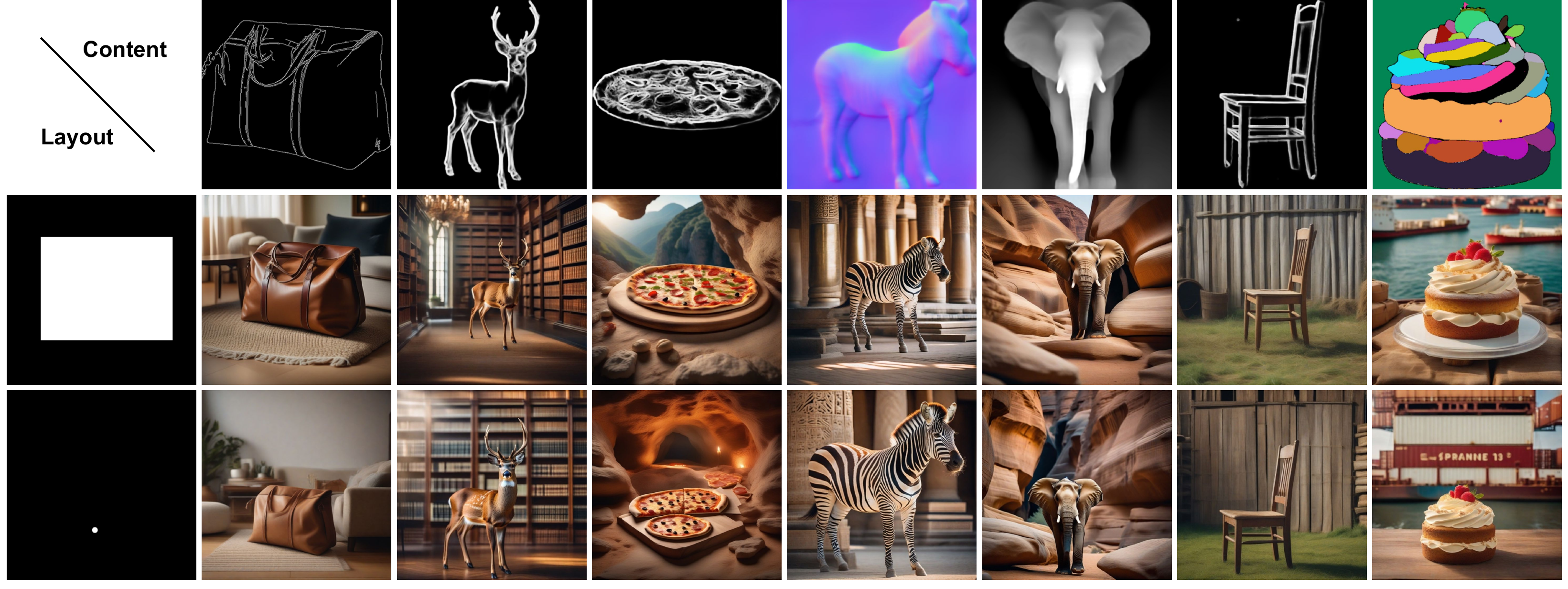}
    \caption{\textbf{The result of only our intra-element controller}. The intra-element controller can effectively and accurately transform the given condition content to the target layout.}
    \label{fig:layout-controller}
    \vspace{-2mm}
\end{figure*}

\begin{figure*}
    \centering
    \includegraphics[width=\linewidth]{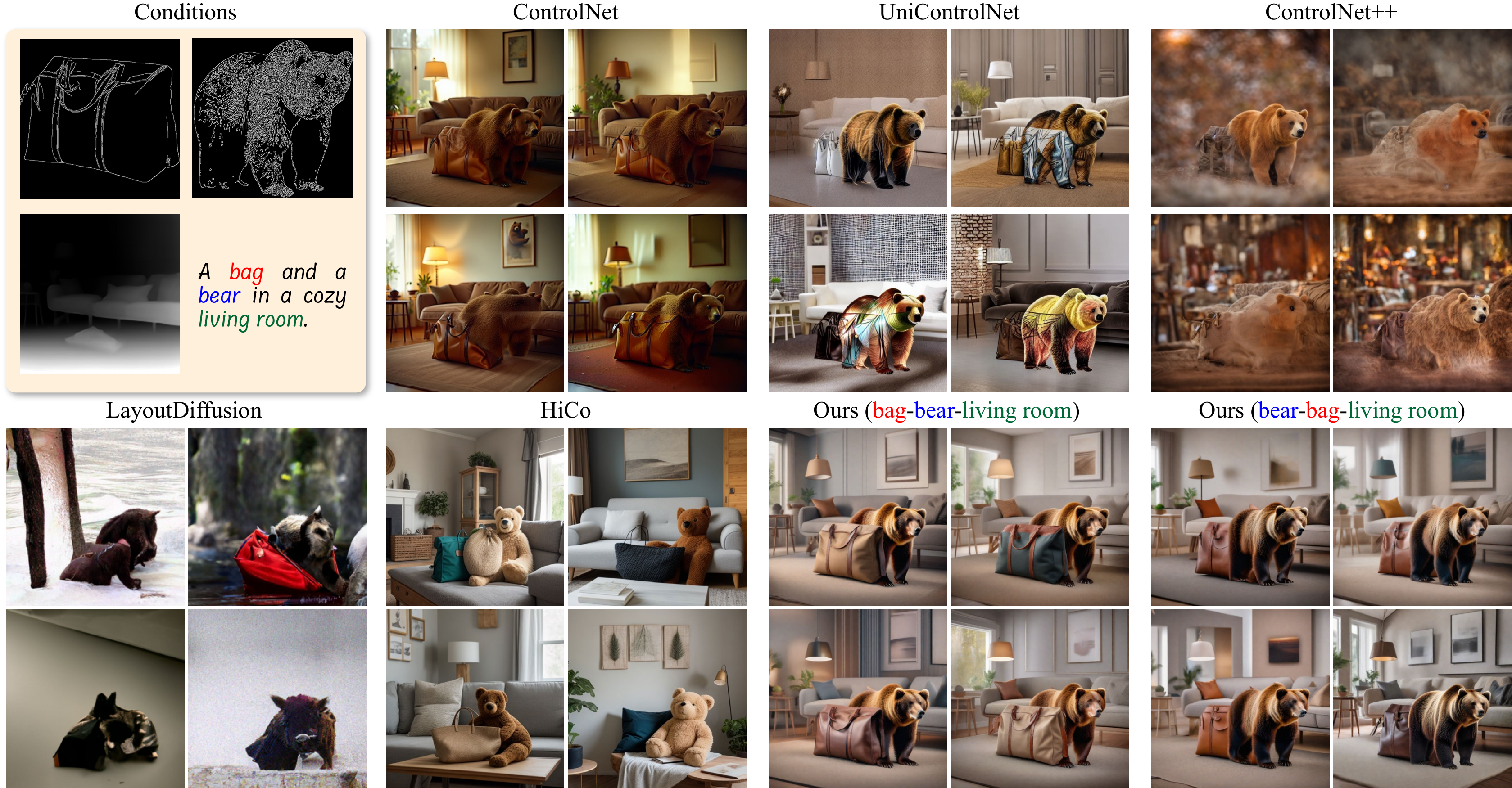}
    \caption{\textbf{Comparisons with ContorlNet~\cite{zhang2023adding}, UniControlNet~\cite{zhao2023uni}, ControlNet++~\cite{li2025controlnet}, Layout Diffusion~\cite{zheng2023layoutdiffusion} and HiCo~\cite{cheng2024hico}.} The prompt we use is ``\textit{A bag and a bear in a cozy living room}''.}
    \label{fig:multi-fusion-result}
    \vspace{-2mm}
\end{figure*}


\subsection{Quatitative Results}
In Fig.~\ref{fig:total_exp}, we show the visual results of our method, it takes the layout condition, content condition of the element, and background as inputs.
Unlike ControlNet, which requires the user to provide detailed layout and content information, our approach decouples content and spatial layout control as separate conditions, allowing the user to control independently. 
Fig.~\ref{fig:layout-controller} illustrates the result of the intra-element controller, our intra-element controller is capable of controlling the ControlNet outputs of different conditions based on varying layout conditions.

To better compare other controllable image generation models, we give a comparison with existing models.
There are two main classes of approaches we compare, one focusing on customizing the content of different condition types and then fusing them {(e.g., ControlNet~\cite{zhang2023adding}, UniControlNet~\cite{zhao2023uni}, and ControlNet++~\cite{li2025controlnet}).} However, this type of approach ignores the fusion strategy when fusing in the conflict areas, leading to obvious artifacts in the conflict region. For example, as shown in Fig.~\ref{fig:multi-fusion-result}, the ``bag'' and ``bear'' melt in the overlap area. 

Another type of approach focuses on generating a specified object within a particular region based on the given prompt. These methods can only specify approximate objects, not {fine-grained control} of content (e.g., Layout Diffusion~\cite{zheng2023layoutdiffusion}, HiCo~\cite{cheng2024hico}). 
Besides, it cannot effectively specify the sequential order of the elements instead relying on prior knowledge from the training datasets. As shown in Fig.~\ref{fig:multi-fusion-result}, the generated content cannot be specified, and the order in which the elements appear is random.

As shown in Fig.~\ref{fig:multi-fusion-result}, our proposed module solves the overlap and conflict problem very well. The corresponding image can be generated by simply adjusting the layer order. For example, if conditions are entered in the order of ``a bag'', ``a bear'' and ``a cozy living room'', an image of a living room with a bag in front of a bear can be generated. Similarly, adjusting the conditional order of ``a bag'' and ``a bear'' produces a different image, with the same elements appearing in a different order.


\subsection{Ablation Studies}

The Inter-Element Controller solves two following problems: unnaturalness when fusing multiple elements and occlusion when fusing multiple elements. 
To represent the layer ordering relationship between different elements, we assign a 1d order embedding to the sorted element feature, enabling the model to perceive the order relationships. As shown in Fig.~\ref{fig:ablation-fusion}, with the order embedding, the model misinterprets the order of the elements, leading to a blending problem similar to traditional ControlNet across different instances.
The same problem arises in the absence of the layer transformer, the model cannot distinguish which element should appear, thereby failing to execute the user's command of positioning a specific element in the foreground. Additionally, the image quality degrades due to the direct mixing of multiple elements.
Additionally, the absence of the spatial transformer may lead to artifacts or unnatural regions in the generated image, which can also be observed in Fig.~\ref{fig:ablation-fusion}.

\begin{figure}[t]
    \centering
    \includegraphics[width=\linewidth]{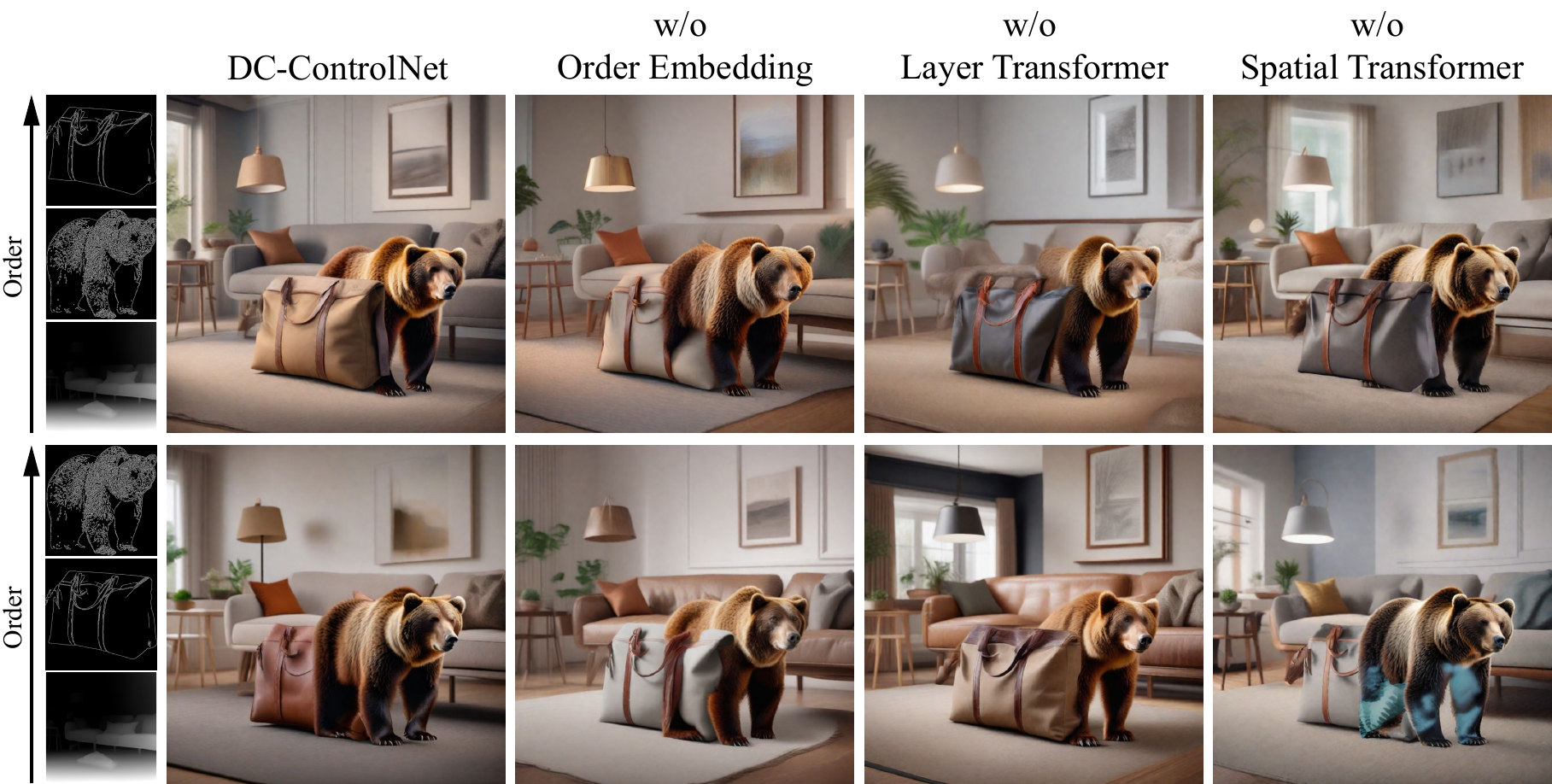}
    \caption{\textbf{The ablation study of Inter-Element Controller}. Order embedding provides the order information between elements, enabling the model to generate occlusion-aware outputs. The absence of the layer transformer and spatial transformer introduces artifacts.}
    \vspace{-1mm}
    \label{fig:ablation-fusion}
\end{figure}

\section{Conclusion}
ControlNet-based generation methods typically rely on global conditions to guide image generation. However, these methods face issues of flexibility and precision in multi-condition image generation tasks, often leading to misunderstandings of conditions and artifacts in the generated output. To address this issue, this paper proposed decoupling the global condition into independent elements, contents, and layouts, introducing DC-ControlNet. Specifically, DC-ControlNet includes an Intra-Element Controller to manage the content and layout conditions within individual elements, ensuring that different conditions within an element are independent and do not interfere with each other. On the other hand, the Inter-Element Controller in DC-ControlNet manages the interactions between different elements, accurately handling occlusions and interaction relationships. Additionally, we presented a new dataset designed for training and evaluating multi-condition generation models. Our experimental results demonstrate that DC-ControlNet outperforms existing methods, providing a more refined and flexible solution for controllable image generation.

{\small
\bibliographystyle{ieee_fullname}
\bibliography{arxiv_dc-ControlNet}
}

\end{document}